\documentclass[10pt,twocolumn,letterpaper]{article}
\usepackage{cvpr}
\usepackage[T1]{fontenc}
\usepackage{amsmath,amssymb,booktabs,array,multirow}
\usepackage{algorithm,algpseudocode}
\usepackage{graphicx,xspace}
\usepackage{placeins}
\usepackage{flafter}
\usepackage{dblfloatfix}
\usepackage{tikz}
\usetikzlibrary{arrows.meta,positioning,fit,backgrounds}
\newcommand{\method}{\textup{\textsc{DMT-Attack}}\xspace}
\newcommand{\pihalf}{\ensuremath{\pi_{0.5}}\xspace}
\newcommand{\CASR}{\ensuremath{\mathrm{cASR}}\xspace}
\newcommand{\SR}{\ensuremath{\mathrm{SR}}\xspace}
\newcommand{\R}{\mathbb{R}}
\DeclareMathOperator*{\argmin}{arg\,min}
\definecolor{cvprblue}{rgb}{0.21,0.49,0.74}
\usepackage[breaklinks,colorlinks,allcolors=cvprblue]{hyperref}
\hypersetup{pdftitle={DMT-Attack: Diffusion-Prior Texture Search for Closed-Loop Vision-Language-Action Policies},pdfauthor={Jun Xia, Shangwen Wang, Ma Jun, Ruichao Ren, Zhouyang Jia, Xin Zhao, Huijun Liu, Jie Yu}}
\def\confName{CVPR}
\def\confYear{2027}
\title{DMT-Attack: Diffusion-Prior Texture Search\\for Closed-Loop Vision-Language-Action Policies}
\author{
Jun Xia\textsuperscript{1}\quad Shangwen Wang\textsuperscript{1}\quad Ma Jun\textsuperscript{1}\quad Ruichao Ren\textsuperscript{1} \\
Zhouyang Jia\textsuperscript{1}\quad Xin Zhao\textsuperscript{1}\quad Huijun Liu\textsuperscript{1}\quad Jie Yu\textsuperscript{1} \\[5pt]
\textsuperscript{1}National University of Defense Technology \\
{\small nudt.edu.cn}
}
\begin{document}
\maketitle
\begin{abstract}
Vision-language-action (VLA) policies rely on visual feedback throughout manipulation, making them vulnerable to persistent texture attacks. However, optimized patterns can stand out from their surroundings, making an attack easier to notice and its carrier more likely to be inspected or removed. To address this limitation, our key insight is to search the structured image distribution of a diffusion model for natural-looking textures while using closed-loop feedback to identify task-disruptive candidates. Based on this idea, we propose \method, a gradient-free gray-box attack that generates textures for ordinary scene objects such as tablecloths. A frozen text-conditioned diffusion model defines the search space, and an eight-expert latent proposal network is initialized from evaluated teacher latents and refined using rollout feedback. We evaluate \method against \pihalf and OpenVLA-OFT across four LIBERO suites. \method achieves a macro clean-conditioned attack success rate (\CASR) of 57.9\%, compared with 44.7\% for UPA-RFAS and 19.7\% for random natural textures. It also attains higher scene similarity and a lower clean-calibrated detector flag rate (45\% versus 97.5--100\% for optimized patch baselines). These results demonstrate that natural-looking textures can disrupt VLA visual feedback and closed-loop decision-making, motivating defenses that jointly assess visual appearance and task behavior.
\end{abstract}

\section{Introduction}
\label{sec:introduction}
\begin{figure}[t]
\centering
\includegraphics[width=\linewidth]{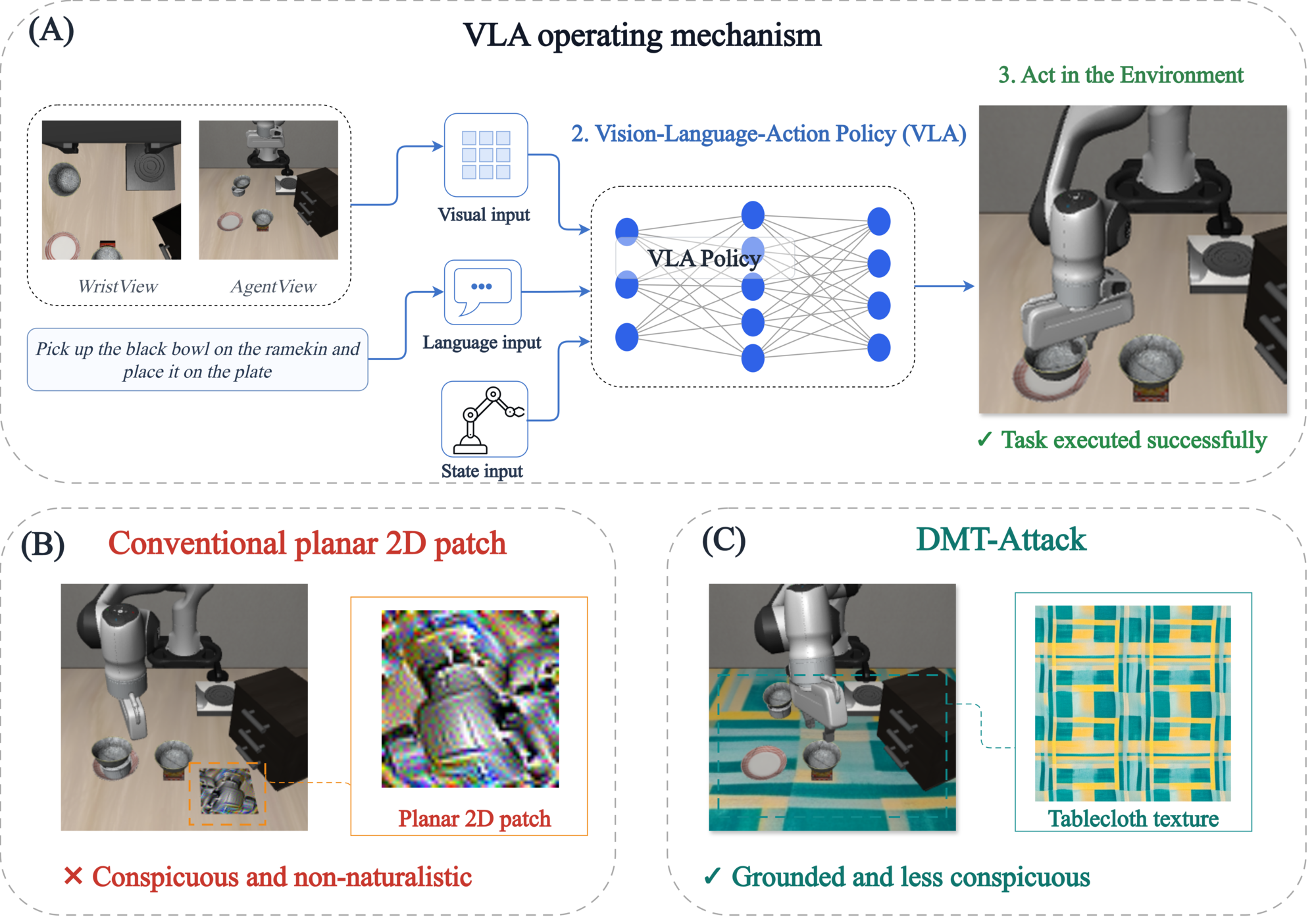}
\caption{\method targets the VLA perception--action loop using persistent textures. (A) The policy uses visual observations, language, and robot state to perform a manipulation task. (B) A conventional patch introduces a conspicuous local pattern. (C) A diffusion prior generates a tabletop texture whose content is searched for task-disruptive effects.}
\label{fig:overview}
\end{figure}

Vision-language-action (VLA) policies combine visual observations, language instructions, and robot state to perform manipulation tasks~\cite{vaswani2017attention,dosovitskiy2021image,ichter2023do,brohan2023rt1,driess2023palme,jiang2023vima,shridhar2023perceiver,chi2023diffusion,brohan2023rt2visionlanguageactionmodelstransfer}. Because these policies repeatedly use visual feedback to choose actions, a persistent texture can affect observations throughout task execution. Evaluating such attacks therefore requires measuring their effects on closed-loop task completion.

Existing texture and patch attacks demonstrate the vulnerability of learned perception and control to visual perturbations~\cite{goodfellow2015explaining,carlini2017towards,madry2018towards,brown2018adversarialpatch,eykholt2018robust,athalye2018synthesizing,karmon2018lavan,thys2019fooling,wang2025exploringadversarialvulnerabilitiesvisionlanguageaction,lu2026robotsobeypatchuniversal}. However, optimized patterns can appear conspicuous against their surroundings~\cite{hu2021naturalistic}. Such visual contrast can make the attack easier for an observer to notice, exposing the modified object as suspicious and prompting its inspection or removal. This motivates searching for textures that disrupt task execution while remaining visually compatible with the scene. To address this limitation, our key insight is that the structured image distributions learned by diffusion models can guide the search toward natural-looking textures, while rollout feedback identifies candidates that disrupt task execution~\cite{sohldickstein2015deep,ho2020denoisingdiffusionprobabilisticmodels,nichol2021improved,song2021score,rombach2022highresolutionimagesynthesislatent,karras2022elucidating,meng2022sdedit}. We use a diffusion prior to generate textures for ordinary scene objects such as tablecloths, allowing the attack carrier to blend into the workspace.

Based on this idea, we propose \method, a diffusion-prior texture attack that searches the latent space of a frozen text-conditioned diffusion model to generate candidate textures. Inspired by the eight harmonic templates used in color harmonization~\cite{cohenor2006color}, we use an eight-expert proposal network whose experts learn diverse latent proposals. The network is initialized by matching previously evaluated teacher latents and refined using closed-loop rollout feedback. An appearance filter screens decoded textures before simulation, while replay and random exploration expand the candidate pool. This gradient-free gray-box search uses source-policy attention and behavioral diagnostics while keeping the VLA policy, diffusion model, and renderer frozen. Together, these components turn naturally generated textures into persistent, task-disruptive scene content through closed-loop feedback.

We evaluate task disruption and visual compatibility using paired task-completion outcomes, CLIP/DINO feature comparisons, maximum mean discrepancy (MMD), and a feature detector calibrated on clean scenes~\cite{radford2021learning,caron2021emerging,gretton2012kernel,liang2023advdm,chen2023diffattack,luo2024visual,gong2023figstep,bailey2024imagehijacks}. Selected textures are tested on held-out initial states. Across \pihalf~\cite{intelligence2025pi05visionlanguageactionmodelopenworld} and OpenVLA-OFT~\cite{kim2025finetuningvisionlanguageactionmodels} on LIBERO-Object, Spatial, Goal, and Long~\cite{liu2023liberobenchmarkingknowledgetransfer}, \method achieves a macro-averaged clean-conditioned attack success rate (\CASR) of 57.9\%, compared with 44.7\% for UPA-RFAS. It attains the highest cASR in all eight policy--suite cells.

In summary, our contributions are:
\begin{enumerate}
    \item We introduce \method, a diffusion-prior texture attack on closed-loop VLA policies that combines a frozen image generator with a learned multi-expert latent proposal network and rollout feedback.
    \item We present an evaluation protocol that assesses task disruption and visual compatibility, with paired task outcomes and testing on initial states excluded from candidate selection.
    \item We evaluate the method across two policies and four LIBERO suites, and examine task-isolated selection, cross-suite transfer, scene appearance, and search-budget effects.
\end{enumerate}

\section{Related Work}
\label{sec:related}
\paragraph{Visual attacks on embodied policies.}
Adversarial examples and patch attacks established threat models for learned visual systems~\cite{goodfellow2015explaining,carlini2017towards,madry2018towards,brown2018adversarialpatch}. Studies of physical transformations and localized perturbations extend these attacks to varied observation conditions and restricted image regions~\cite{eykholt2018robust,athalye2018synthesizing,karmon2018lavan,thys2019fooling,zhang2022adversarialpatchmultiscale}. Naturalistic patch generation uses natural-image priors to address visual conspicuousness~\cite{hu2021naturalistic}. In VLA robotics, recent work examines physical vulnerabilities, action deviation, representation misalignment, model-agnostic transfer, and universal patches~\cite{cheng2025manipulationfacingthreatsevaluating,wang2025exploringadversarialvulnerabilitiesvisionlanguageaction,xu2025modelagnosticadversarialattackdefense,lu2026robotsobeypatchuniversal}. Other attacks target reasoning pathways, visual proprioception, and partial observability~\cite{huang2026traphijackingvlacotreasoning,fu2026vlahijacktransferablepatchattack,wang2026partiallyobservableadversarialpatch}. Our focus is the joint evaluation of task disruption and visual compatibility: \method searches a diffusion-generated texture family and compares the resulting textures with optimized patch baselines under both criteria.

\paragraph{Diffusion priors and proposal learning.}
Denoising, score-based, and latent diffusion models provide structured image generators with controllable sampling trajectories~\cite{sohldickstein2015deep,ho2020denoisingdiffusionprobabilisticmodels,nichol2021improved,song2021score,rombach2022highresolutionimagesynthesislatent,karras2022elucidating,meng2022sdedit}. Adversarial research also studies generative models as attack targets and purification defenses~\cite{liang2023advdm,chen2023diffattack}. \method uses a frozen diffusion model to define a texture search space and learns a multi-expert proposal from rollout feedback. This formulation combines a generative image prior with optimization against closed-loop task outcomes.

\paragraph{VLA foundations and embodied benchmarks.}
Transformer attention and vision transformers underpin modern multimodal policies~\cite{vaswani2017attention,dosovitskiy2021image}. Embodied models such as RT-1, PaLM-E, VIMA, Perceiver-Actor, and Diffusion Policy connect visual or multimodal observations to robot actions~\cite{brohan2023rt1,driess2023palme,jiang2023vima,shridhar2023perceiver,chi2023diffusion,ichter2023do}. LIBERO benchmarks knowledge transfer across manipulation tasks~\cite{liu2023liberobenchmarkingknowledgetransfer}. We use its task-completion predicates to evaluate persistent texture attacks.

\paragraph{Appearance-aware evaluation.}
CLIP and DINO provide feature spaces for image comparison~\cite{radford2021learning,caron2021emerging}, while MMD summarizes differences between image collections~\cite{gretton2012kernel}. Natural-image benchmarks and imperceptible-attack studies provide context for evaluating appearance alongside attack strength~\cite{hendrycks2019benchmarking,chen2023imperceptible,croce2020reliable,hendrycks2021natural}. Related visual-generation and image-hijacking work examines how image content can alter model behavior~\cite{luo2024visual,gong2023figstep,bailey2024imagehijacks}. We report feature similarity and anomaly-detector flag rates alongside overall task success and failures among paired clean-successful episodes.

\section{Diffusion-prior Texture Search}
\label{sec:method}
\subsection{Threat model and objective}
Let a frozen policy $\pi$ produce actions from observations $o_t$ and a benign instruction $\ell$. The attacker controls one static $512\times512$ texture. The attack uses an ordinary scene object, such as a tablecloth, as its carrier. The generated natural texture blends the carrier into the workspace while persistently influencing the policy's visual observations throughout an episode. The instruction, controller, and task-completion objective remain unchanged.

Search uses paired clean/patched behavioral traces and attention-derived diagnostics from the source policy. This \emph{gradient-free gray-box search} updates proposal-side parameters through rollout rewards while keeping the VLA, diffusion model, and renderer frozen. Once selected, the texture remains fixed on the scene carrier during policy execution.

Algorithm~\ref{alg:dmt} summarizes the complete search pipeline. Phase I transfers evaluated high-gap latents into the multi-expert proposal family. Phase II alternates candidate generation, appearance screening, closed-loop feedback, and proposal updates; the final texture is selected only after online refinement and then locked for held-out evaluation.

\begin{algorithm}[t]
\caption{DMT-Attack Search}
\label{alg:dmt}
\begin{algorithmic}[1]
\Require Frozen policy $\pi$ and decoder $G$; prompt $p$; teachers $\mathcal{T}$; appearance filter $F$; online budget $N$; validation/test episodes $\mathcal{E}_{\rm val},\mathcal{E}_{\rm test}$
\Ensure Locked texture $x^*$ and test results
\State Initialize proposal $\theta$, critic $V_\psi$, exploration variances $\{\sigma_h\}$, and buffers $\mathcal{B},\mathcal{C}\gets\emptyset$
\Statex \textbf{Phase I: Teacher initialization}
\ForAll{$(p_i,z_{T,i}^*,g_i)\in\mathcal{T}$}
    \State Compute expert latents and costs $\{z_i^h,c_i^h\}$
    \State $h_i^*\gets\argmin_h c_i^h$ \Comment{\eqref{eq:teacher}}
    \State Update $\theta$ by minimizing $L_{\rm I}$ in \eqref{eq:offline}
    \State Add high-gap teacher latents to $\mathcal{B}$
\EndFor
\Statex \textbf{Phase II: Closed-loop refinement}
\For{$t=1,\ldots,N$}
    \State Sample three model proposals
    \State Add replay and random-exploration proposals
    \ForAll{candidate latents $z_T$}
        \State Project $z_T$ to $\widetilde z_T$ using \eqref{eq:projection}
        \State $x\gets G(\widetilde z_T,p)$
        \If{$F(x)=1$}
            \State Render $x$ on the scene carrier
            \State Collect paired rollouts with frozen $\pi$
            \State Compute $g_{\rm train}$ and $R(u)$ using \eqref{eq:curriculum}
            \State $\mathcal{C}\gets\mathcal{C}\cup\{x\}$
        \EndIf
    \EndFor
    \State Update $\theta$, $V_\psi$, and $\{\sigma_h\}$ by PPO on model-proposal rewards
    \State Add eligible high-gap latents to $\mathcal{B}$
\EndFor
\State Select $x^*$ from $\mathcal{C}$ using validation on $\mathcal{E}_{\rm val}$
\State Lock $x^*$; evaluate on held-out episodes $\mathcal{E}_{\rm test}$
\State \Return $x^*$ and test results
\end{algorithmic}
\end{algorithm}

For episode $i$, let $C_i,A_i\in\{0,1\}$ indicate clean and attacked task success. The attacked task success rate (\SR) and clean-conditioned attack success rate are
\begin{equation}
\SR=\frac{1}{N}\sum_i A_i,\qquad
\CASR=\frac{\sum_i C_i(1-A_i)}{\sum_i C_i}.
\label{eq:metrics}
\end{equation}
Higher \CASR indicates greater disruption of episodes the clean policy solves, while \SR measures overall attacked execution performance. Reporting both metrics distinguishes disruption of clean-successful episodes from overall task performance, which also includes clean-failed episodes that succeed under the changed texture.

\begin{figure*}[t]
\centering
\includegraphics[width=\textwidth]{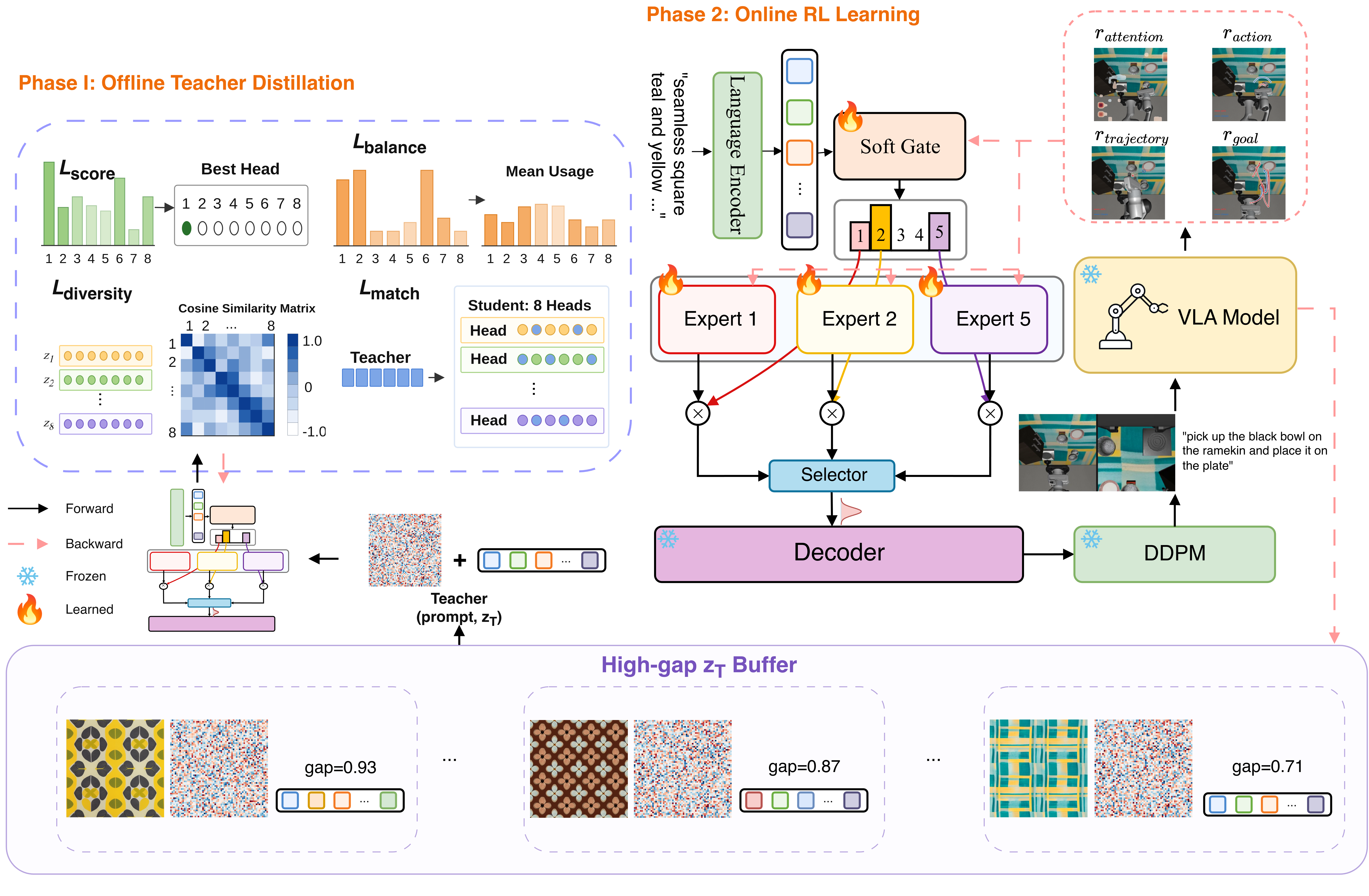}
\caption{Main architecture and learning procedure. Phase I distills high-gap teacher latents into an eight-expert text-conditioned proposal family using matching, diversity, and load-balancing objectives. In Phase II, a frozen diffusion model decodes the selected latent into a candidate texture; appearance-screened candidates are evaluated by a frozen VLA in closed-loop rollouts, and paired behavioral signals update the proposal, critic, and exploration policy. The diffusion model, renderer, and VLA remain frozen.}
\label{fig:method}
\end{figure*}

\subsection{Text-conditioned multi-expert proposals}
\cref{fig:method} shows the two-stage learning framework. Its ``Soft Gate'' routes expert proposals, and the appearance filter screens decoded textures. Dashed feedback paths carry rollout rewards to the trainable proposal and optimization modules.
The frozen Stable Diffusion 2.1 text encoder supplies 77 token embeddings and a pooled 1024-dimensional end-of-sequence representation. A shared $1024\!\rightarrow\!512\!\rightarrow\!512\!\rightarrow\!512$ backbone uses LayerNorm and SiLU. Eight independent heads produce 128-dimensional bottlenecks; a text-conditioned decoder maps each into $z_T^h\in\R^{4\times64\times64}$. Its upsampling blocks use token cross-attention and pooled-conditioning modulation. A soft routing network produces expert weights $q_h$, and a separate score head chooses a preferred proposal with two companion heads for exploration. We compute all latent heads and decode the candidates selected for evaluation.

Each initial diffusion latent is projected to a fixed radius,
\begin{equation}
\widetilde z_T=\sqrt{D}\,z_T/\max(\|z_T\|_2,\epsilon),\quad D=16384.
\label{eq:projection}
\end{equation}
This controls the latent norm near the typical Gaussian radius. The frozen diffusion model maps $(\widetilde z_T,p)$ to a texture using 30 sampling steps and classifier-free guidance 7.5. Search optimizes the initial diffusion latent with fixed reverse-process parameters.

\subsection{Phase I: teacher initialization}
Teacher tuples $(p_i,z_{T,i}^{*},g_i)$ contain a prompt, a previously evaluated latent, and its task-gap record. For each tuple, the closest expert minimizes
\begin{equation}
\begin{split}
c_i^h&=\operatorname{MSE}(z_i^h,z_{T,i}^{*})
       +0.3\bigl(1-\cos(z_i^h,z_{T,i}^{*})\bigr),\\
h_i^*&=\argmin_h c_i^h.
\end{split}
\label{eq:teacher}
\end{equation}
The winning expert receives matching supervision; the score logits $s_i$ receive cross-entropy supervision for $h_i^*$. The score head therefore ranks experts by teacher-latent matching. With cosine matrix $K_i$ over normalized expert latents, diversity and balance regularizers give
\begin{equation}
\begin{split}
L_{\rm div}&=-\log\det(K_i+0.1I),\quad
L_{\rm bal}=8\sum_h\bar q_h^2,\\
L_{\rm I}&=c_i^{h_i^*}+0.1\operatorname{CE}(s_i,h_i^*)
                 +0.01L_{\rm div}+0.05L_{\rm bal}.
\end{split}
\label{eq:offline}
\end{equation}
High-gap teachers receive additional replay. In the budget experiment (\cref{sec:phase1}), teacher collection uses 200 search rollouts on tasks 0--4, followed by evaluation on disjoint task/state cells. We account for this collection cost separately from online refinement.

\subsection{Phase II: closed-loop refinement}
\paragraph{Appearance filter.}
A decoded texture is screened using total-variation ratio, high-frequency energy, the 95th saturation percentile, and a lower bound on color entropy. This low-level filter screens candidates before simulation: each rejected candidate incurs one diffusion decode and zero rollouts. Expert routing operates upstream on latent proposals.

\paragraph{Exploration and replay.}
The proposal policy samples in the 128-dimensional expert bottleneck. Each prompt produces three model candidates, with a replay-latent perturbation considered with probability 0.7 and a random exploration candidate with probability 0.2. Eligible high-gap latents ($g_{\rm train}>0.2$) enter a buffer of at most 30 entries. A fresh run can use a 300-step buffer-matching warm start; buffer matching is also added to half the online updates with weight 0.05. Replay and random exploration expand the candidate pool, and proximal policy optimization (PPO) updates the sampled model proposals.

\paragraph{Feedback curriculum.}
The training gap is $g_{\rm train}=1-\SR_{\rm rollout}$; final reporting uses \cref{eq:metrics}. Let $u\in[0,1]$ denote online progress. The curriculum is
\begin{equation}
R(u)=\begin{cases}
6r_{\rm att}, &u<0.25,\\
g_{\rm train}+6r_{\rm att}, &0.25\leq u<0.55,\\
g_{\rm train}+6r_{\rm att}+r_{\rm beh}, &u\geq0.55,
\end{cases}
\label{eq:curriculum}
\end{equation}
where $r_{\rm beh}=1.5r_{\rm act}+6r_{\rm traj}+3r_{\rm goal}$. All four signals are computed against paired clean traces. The attention term compares agent- and wrist-view saliency maps. The action term averages per-step action differences normalized by the clean action norm; the trajectory term measures mean Euclidean end-effector displacement, and the goal term compares mean end-effector positions over the final five steps of each rollout.

The schedule progressively connects visual response to task outcome and execution behavior. Early attention displacement supplies dense guidance when terminal failures are sparse. The middle stage adds $g_{\rm train}$ to favor task-disruptive candidates. The final stage incorporates immediate action changes and accumulated spatial deviations to refine their behavioral effects. PPO updates the proposal, critic, and per-expert exploration variances, while the diffusion model and VLA remain frozen.

\paragraph{Candidate selection and testing.}
Online training produces a candidate pool, validation selects the final texture, and testing evaluates the locked texture on episodes excluded from search and selection. Teacher collection, optimization, diffusion decoding, and training rollouts constitute upstream construction costs; the budget ablation examines teacher initialization and online search duration.

\section{Experiments}
\label{sec:experiments}

\begin{table*}[t]
  \centering
  \small
  \setlength{\tabcolsep}{2.5pt}
  \providecommand{\MainResultsCaption}{Independent-state evaluation under the same candidate-selection protocol for both policies and all attack methods. Per method, policy, and suite, four candidates receive ten validation rollouts at state 20; the selected texture is locked before testing ten tasks at states 30--32 ($N=30$). Placement is controlled in simulation using the same upper-left image region for all attacks. Cells are SR/cASR (\%). Macro cASR equally weights the eight model--suite cells. Approximate 95\% intervals follow the conditional-binomial model in \cref{eq:ci}. Bold marks the highest cASR.}
  \caption{\MainResultsCaption}
  \label{tab:main}
  \resizebox{\textwidth}{!}{%
  \begin{tabular}{l cccc cccc cc}
    \toprule
    & \multicolumn{4}{c}{$\pi_{0.5}$~\cite{intelligence2025pi05visionlanguageactionmodelopenworld}} & \multicolumn{4}{c}{OpenVLA-OFT~\cite{kim2025finetuningvisionlanguageactionmodels}} & & \\
    \cmidrule(lr){2-5}\cmidrule(lr){6-9}
    Method & Object & Spatial & Goal & Long
           & Object & Spatial & Goal & Long
           & Macro & 95\% CI \\
    \midrule
    Clean                  & 86.7/-- & 90.0/-- & 96.7/-- & 90.0/-- & 60.0/-- & 90.0/-- & 73.3/-- & 86.7/-- & -- & -- \\
    Random natural         & 86.7/11.5 & 90.0/3.7 & 100.0/0.0 & 83.3/14.8 & 66.7/33.3 & 73.3/22.2 & 60.0/40.9 & 73.3/30.8 & 19.7 & [14.3, 25.0] \\
    UADA                   & 73.3/23.1 & 73.3/22.2 & 93.3/6.9 & 46.7/48.1 & 13.3/88.9 & 43.3/51.9 & 46.7/50.0 & 60.0/30.8 & 40.2 & [34.3, 46.2] \\
    UPA                    & 43.3/50.0 & 76.7/18.5 & 96.7/3.4 & 56.7/44.4 & 3.3/94.4 & 63.3/29.6 & 36.7/59.1 & 56.7/34.6 & 41.8 & [36.0, 47.6] \\
    UPA-RFAS               & 40.0/53.8 & 73.3/22.2 & 100.0/0.0 & 53.3/40.7 & 13.3/88.9 & 43.3/55.6 & 43.3/50.0 & 53.3/46.2 & 44.7 & [38.7, 50.7] \\
    \method                & 26.7/\textbf{69.2} & 63.3/\textbf{33.3} & 80.0/\textbf{20.7} & 40.0/\textbf{59.3} & 0.0/\textbf{100.0}  & 36.7/\textbf{59.3} & 36.7/\textbf{63.6} & 36.7/\textbf{57.7} & \textbf{57.9} & [51.9, 63.9] \\
    \bottomrule
  \end{tabular}}
\end{table*}

\begin{figure}[!t]
\centering
\includegraphics[width=\linewidth]{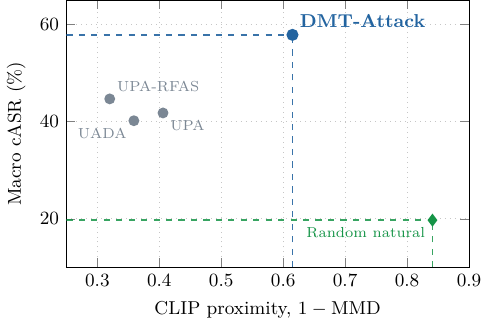}
\caption{Method-level attack and appearance summary. The vertical coordinate reports macro cASR from the independent-state main experiment; the horizontal coordinate reports $1-\mathrm{MMD}$ from the separate CLIP texture-collection evaluation.}
\label{fig:tradeoff}
\end{figure}

We evaluate attack effectiveness across policies and suites, task and suite generalization, and scene appearance, followed by ablations of search parameterization, budget, and teacher initialization. Each experiment specifies its task and state split.

\subsection{Evaluation setup and statistical units}
\label{sec:setup}
\paragraph{Policies, tasks, and carrier.}
The main experiment uses frozen \pihalf~\cite{intelligence2025pi05visionlanguageactionmodelopenworld} and OpenVLA-OFT~\cite{kim2025finetuningvisionlanguageactionmodels} policies in MuJoCo on LIBERO-Object, Spatial, Goal, and Long. Each suite contains ten tasks. To control placement in simulation, all attack methods use the same upper-left image region throughout validation and testing. Camera configuration, instructions, controllers, and success predicates remain fixed within each comparison. We compare clean execution, random natural textures, UADA, UPA, UPA-RFAS, and \method. We evaluate the UADA, UPA, and UPA-RFAS baseline configurations~\cite{wang2025exploringadversarialvulnerabilitiesvisionlanguageaction,lu2026robotsobeypatchuniversal} under the common selection and testing protocol. Random natural textures test a non-task-optimized texture source.

\paragraph{Clean reference.}
For task evaluation, a clean scene uses the solid-color reference image (black, white, or red) specified by the experimental configuration in the controlled image region. This reference preserves the image-placement procedure and experimental settings used for attack evaluation, controlling for their effects when comparing texture content. Conditioning \CASR on successful reference episodes further excludes failures already present under this configuration. The Clean row in \cref{tab:main} reports task success under this reference. The corresponding episode-level success indicator $C_i$ in \cref{eq:metrics} determines which episodes enter the clean-conditioned \CASR calculation.

\paragraph{Equal validation selection and independent testing.}
For each policy, suite, and attack method, four candidates each receive ten validation rollouts, one per task, at initial-state index 20. The best validation candidate is then frozen. All choices are locked before the final paired tests on the same ten tasks at initial-state indices 30, 31, and 32. Thus, each model--suite test cell has 30 episodes, and each method has 240 attacked test episodes across the two policies. The validation cost is 40 candidate rollouts per model--suite cell, or 320 per method across eight cells, excluding clean-reference runs and upstream candidate construction. Validation and testing share task identities and use disjoint initial-state indices.

\paragraph{Endpoints and uncertainty.}
\cref{tab:main} reports attacked \SR and paired-clean \CASR. The eight clean-success denominators, in table order, are $(26,27,29,27,18,27,22,26)$. Macro \CASR assigns equal weight to the eight cells. For cell estimates $\widehat p_j$ and clean-success counts $n_j$, we compute approximate intervals as
\begin{equation}
\bar p=\frac18\sum_{j=1}^{8}\widehat p_j,\qquad
\widehat{\mathrm{SE}}(\bar p)=\frac18
\sqrt{\sum_{j=1}^{8}\frac{\widehat p_j(1-\widehat p_j)}{n_j}}.
\label{eq:ci}
\end{equation}
The approximate 95\% interval is $\bar p\pm1.96\widehat{\mathrm{SE}}$, using independent conditional-binomial cell variances for the fixed candidates. The interval quantifies cell-level sampling uncertainty conditional on the selected textures and the independent-binomial assumption. The following auxiliary experiments report point estimates, with interval bounds also shown in the search-budget plot.

\subsection{Attack effectiveness across policies and suites}
\label{sec:main_results}
\paragraph{Aggregate effect.}
\method achieves 57.9\% macro \CASR, compared with 44.7\% for UPA-RFAS, 41.8\% for UPA, 40.2\% for UADA, and 19.7\% for random natural textures. The gain is 13.2 percentage points over the strongest aggregate baseline and 38.2 points over the random-natural reference. Each selected artifact is evaluated on unseen initial states, so the comparison measures performance beyond the validation episodes.

The corresponding mean \SR across cells falls from 84.2\% for clean execution to 40.0\% under \method; UPA-RFAS yields 52.5\%. Together, \SR and \CASR show reduced overall task completion and disruption of tasks the clean policy solves.

\paragraph{Policy and suite variation.}
Averaging the four suite \CASR values within each policy gives 45.6\% for \pihalf and 70.2\% for OpenVLA-OFT under \method, versus 29.2\% and 60.2\% for UPA-RFAS. \method achieves the highest cASR in all eight cells. OpenVLA-OFT-Object reaches 100\% \CASR on its 18 clean-successful episodes, while \pihalf-Goal reaches 20.7\% on 29 such episodes. Attack effectiveness thus varies across policies and suites, with \pihalf-Goal showing the lowest DMT cASR.

\paragraph{Effect of task-guided texture search.}
The 38.2-point gain over random natural textures demonstrates the value of task-guided texture search. Teacher initialization, replay, and rollout feedback direct the search toward task-disruptive textures.

\subsection{Task and suite generalization}
\label{sec:generalization}
The task-isolated \pihalf experiment selects on tasks 0--4 at state 20 and tests on tasks 5--9 at states 30--32. It retains four candidates per method and freezes the selected artifact before testing. \method reaches 34.7\% macro \CASR, followed by UADA at 29.7\%. Random RGB is the control in this experiment.

\begin{table}[!htbp]
\centering\small
\caption{Task-isolated \pihalf experiment: select on tasks 0--4 and test on tasks 5--9. Values are macro cASR (\%).}
\label{tab:tasks}
\begin{tabular}{lc}
\toprule
Method & Macro cASR (\%)\\
\midrule
Random RGB & 1.7\\
UADA & 29.7\\
UPA & 20.5\\
UPA-RFAS & 27.7\\
\method & \textbf{34.7}\\
\bottomrule
\end{tabular}
\end{table}

\paragraph{Cross-suite selection transfer.}
\cref{tab:transfer} evaluates DMT textures selected on a source suite and frozen for each target suite. Off-diagonal directions average 37.5\% \CASR, while the diagonal averages 35.7\%. Source-Goal to target-Long reaches 45.8\%, whereas source-Object to target-Goal reaches 14.8\%.

\begin{figure*}[!t]
\centering
\includegraphics[width=0.80\textwidth]{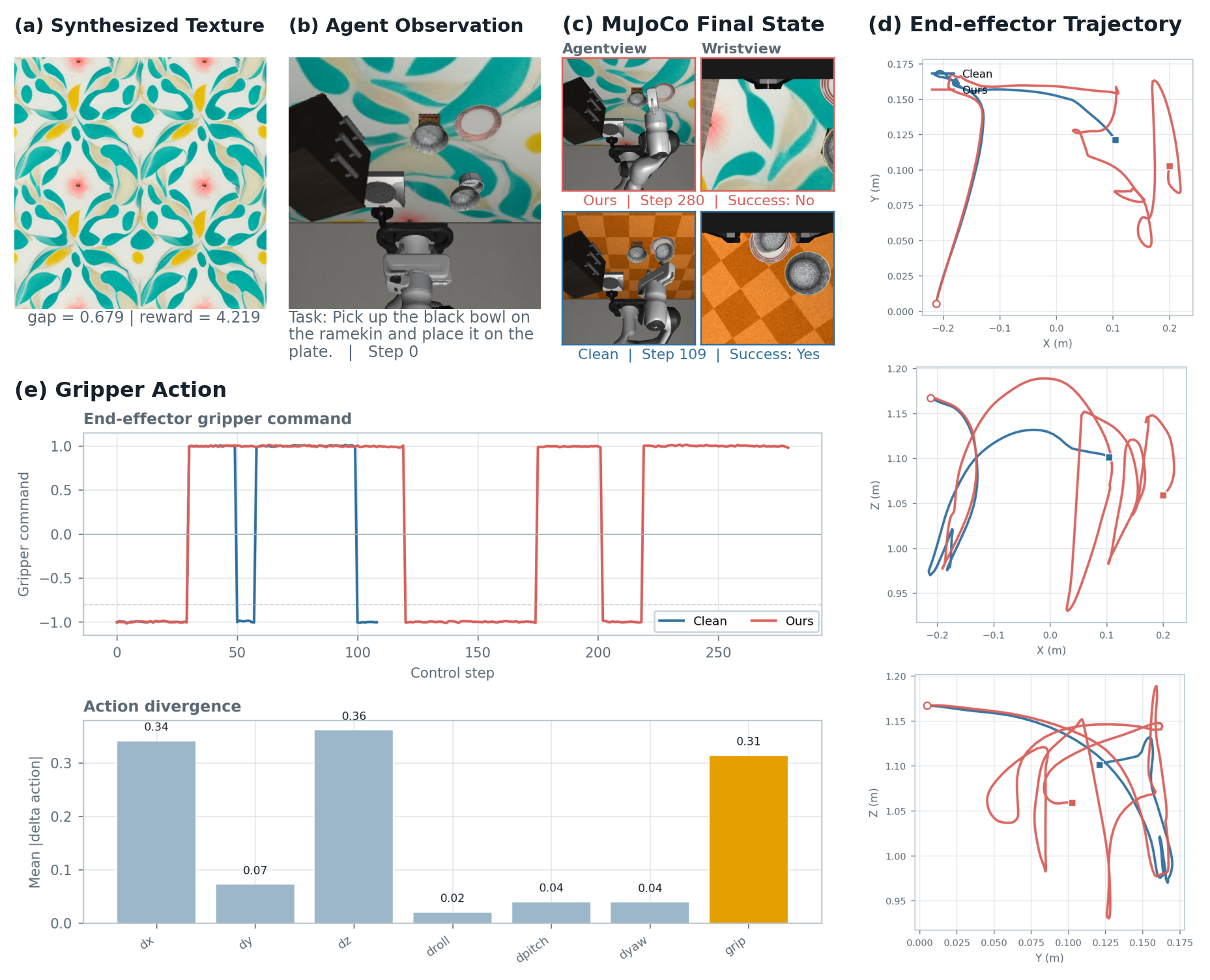}
\caption{Qualitative paired closed-loop trace. The generated texture changes the agent and wrist observations, produces a divergent end-effector path and gripper schedule, and fails at step 280; the paired clean rollout succeeds at step 109. Panels show the texture, observations, end-effector paths, and gripper commands.}
\label{fig:eef}
\end{figure*}

The transfer matrix reveals consistent target-suite differences: across all four sources, \CASR ranges from 14.8--29.6\% on Goal and 38.1--52.4\% on Spatial. Goal's lower susceptibility agrees with the main \pihalf results. Source suites yield complementary textures: Spatial-selected textures perform best on Object, Object-selected textures on Spatial, and Goal-selected textures on Long. These directional patterns and the 37.5\% off-diagonal average demonstrate cross-suite attack capability, with effectiveness varying by both texture source and target suite.

\begin{table}[!htbp]
\centering\small
\caption{Cross-suite DMT selection transfer on \pihalf (cASR, \%). Rows specify the selection suite; columns specify the test suite.}
\label{tab:transfer}
\begin{tabular}{lrrrr}
\toprule
Source $\backslash$ target & Object & Spatial & Goal & Long\\
\midrule
Object & 44.4 & 52.4 & 14.8 & 37.5\\
Spatial & 48.1 & 42.9 & 29.6 & 37.5\\
Goal & 40.7 & 42.9 & 22.2 & 45.8\\
Long & 40.7 & 38.1 & 22.2 & 33.3\\
\bottomrule
\end{tabular}
\end{table}

\subsection{Scene appearance and anomaly detection}
\label{sec:appearance}

\paragraph{Scene-matched feature detector.}
The appearance experiment covers 40 LIBERO tasks, three additional states (47--49), and two camera views. To control scene conditions, all methods share the upper-left image placement, fixed lighting, and fixed cameras. CLIP scene similarity compares attacked and corresponding clean scenes. Detection uses full-agentview features from frozen CLIP ViT-B/32, with a variance-shrunk diagonal Mahalanobis score. Per-dimension clean means and variances are estimated on a fitting set, and the threshold is the 95th score percentile on a separate clean calibration set. Both stages use clean images.

\begin{table}[!htbp]
\centering\small
\caption{Scene similarity and anomaly detection across 40 LIBERO tasks. CLIP similarity compares matched clean and attacked scenes; flag rates and mean anomaly scores use full-agentview features from a clean-calibrated detector.}
\label{tab:detector}
\begin{tabular}{lccc}
\toprule
Method & CLIP $\uparrow$ & Flag rate $\downarrow$ & Score $\downarrow$\\
\midrule
UADA & 0.8554 & 97.5\% & 2.419\\
UPA & 0.8616 & 100.0\% & 2.365\\
UPA-RFAS & 0.8566 & 100.0\% & 2.262\\
\method & \textbf{0.8957} & \textbf{45.0\%} & \textbf{1.836}\\
\bottomrule
\end{tabular}
\end{table}

\method achieves the highest scene similarity (0.8957), lowest flag rate (45\%), and lowest mean anomaly score (1.836) among the evaluated attacks (\cref{tab:detector}). Its flag rate is 52.5 percentage points below UADA and 55 points below UPA and UPA-RFAS. The higher feature similarity and lower detector scores demonstrate improved visual compatibility with clean scenes.

\paragraph{Collection-level proximity.}
\cref{tab:distribution} reports the texture-collection analysis: DMT versus UPA-RFAS yields CLIP nearest-neighbor similarity of 0.725 versus 0.591, CLIP MMD of 0.385 versus 0.680, and DINO MMD of 0.618 versus 1.263. Figure~\ref{fig:tradeoff} summarizes methods using the main-table macro cASR and the separately evaluated texture-collection proximity.

\paragraph{Qualitative closed-loop trace.}
Figure~\ref{fig:eef} shows a paired pick-and-place example. Clean execution succeeds at step 109, whereas the textured rollout remains unsuccessful at step 280. End-effector paths diverge across the three coordinate projections and gripper commands differ over time. The paired trace shows a persistent texture disrupting closed-loop execution through altered actions and a failed manipulation trajectory.

\begin{table}[!htbp]
\centering\small
\caption{Texture-collection proximity analysis. CLIP NN is nearest-neighbor similarity; CLIP and DINO MMD use COCO and LIBERO reference collections, respectively.}
\label{tab:distribution}
\setlength{\tabcolsep}{2.5pt}
\resizebox{\columnwidth}{!}{%
\begin{tabular}{lccc}
\toprule
Group & CLIP NN $\uparrow$ & CLIP MMD $\downarrow$ & DINO MMD $\downarrow$\\
\midrule
\multicolumn{4}{l}{\textit{Optimized attack methods}}\\
\method & \textbf{0.725} & \textbf{0.385} & \textbf{0.618}\\
UADA & 0.622 & 0.641 & 1.270\\
UPA & 0.706 & 0.594 & 1.267\\
UPA-RFAS & 0.591 & 0.680 & 1.263\\
\midrule
\multicolumn{4}{l}{\textit{Texture references}}\\
Random RGB & 0.602 & 0.670 & 1.301\\
Random natural texture & 0.772 & 0.159 & 0.451\\
\bottomrule
\end{tabular}}
\end{table}

\subsection{Ablation studies}
\label{sec:ablations}
\label{sec:phase1}
We examine the search parameterization and teacher initialization through complementary budget ablations.

\begin{figure}[!htbp]
\centering
\includegraphics[width=\linewidth]{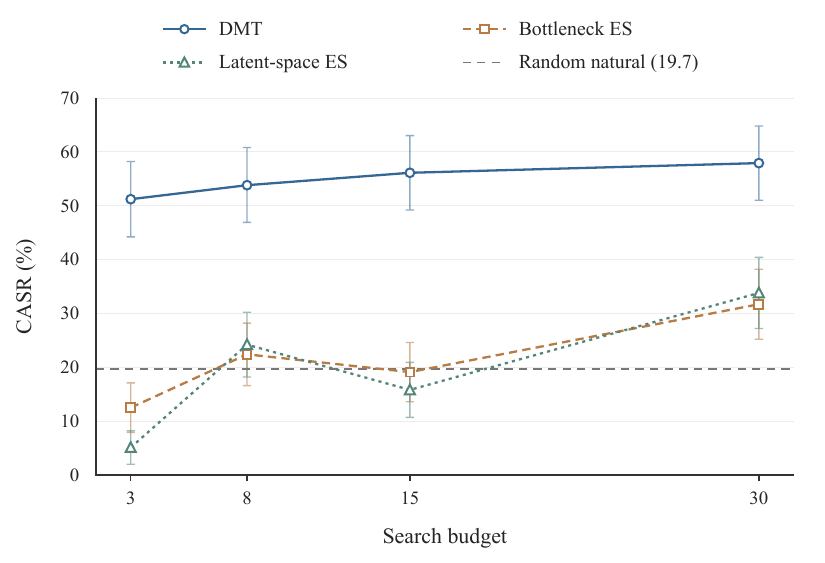}
\caption{Search-budget ablation comparing DMT, Bottleneck ES, and Latent-space ES. Error bars indicate the intervals around cASR estimates. The dashed line marks random natural textures at 19.7\% cASR.}
\label{fig:budget_ablation}
\end{figure}

\paragraph{Search parameterization and budget.}
\cref{fig:budget_ablation} compares \method with two evolution strategies (ES) variants, Bottleneck ES and Latent-space ES, at search budgets of 3, 8, 15, and 30. DMT cASR increases from 51.2\% to 57.9\%, exceeding both ES variants at every evaluated budget. At budget 30, Bottleneck ES and Latent-space ES reach 31.7\% and 33.8\%, respectively; DMT leads the stronger ES variant by 24.1 percentage points. At budget 3, DMT already reaches 51.2\%, above both ES variants at budget 30. Learned proposals deliver strong attacks at small search budgets and sustain their advantage as the budget increases.

\paragraph{Teacher initialization.}
The teacher-initialization ablation uses three paired seeds, selects on task 0 at state 43, and tests on tasks 5--9 at states 45--46. Its teacher tuples come from a prior black-box search on tasks 0--4 with 200 rollouts, disjoint from these test cells. The frozen teacher attains 12.8\% \CASR. \cref{tab:budget} compares full initialization and removal of Phase I at three and 24 online rounds.

\begin{table}[!htbp]
\centering\small
\caption{Phase I budget ablation. $B$ counts online rounds; time reports the duration of online search. Teacher collection uses an additional 200 rollouts.}
\label{tab:budget}
\setlength{\tabcolsep}{3pt}
\begin{tabular}{clcc}
\toprule
$B$ & Variant & cASR (\%) & Time (s)\\
\midrule
3 & Full & 33.3 & 101.7\\
3 & w/o Phase I & 0.0 & 83.6\\
24 & Full & 37.0 & 604.9\\
24 & w/o Phase I & 29.6 & 592.7\\
\bottomrule
\end{tabular}
\end{table}

At three rounds, teacher initialization increases \CASR from 0.0\% to 33.3\%. At 24 rounds, the uninitialized variant reaches 29.6\%, compared with 37.0\% for the full method. Thus, Phase I improves short-budget search, with the largest gain at three online rounds. Full search takes 101.7 s at three rounds and 604.9 s at 24 rounds; the corresponding durations without Phase I are 83.6 s and 592.7 s.

Of the selected textures, 62\% originate from model proposals, 26\% from replay, and 12\% from random exploration. Model proposals supply most selected textures, complemented by replay and random exploration.

\section{Conclusion}
\label{sec:conclusion}
\method uses a frozen diffusion generator to parameterize persistent workspace textures and closed-loop feedback to search for task-disruptive candidates. Across two policies and four LIBERO suites, it achieves 57.9\% macro cASR, exceeding UPA-RFAS by 13.2 percentage points. Appearance experiments show higher scene similarity and a 45\% detector flag rate, compared with 97.5--100\% for optimized patch baselines. Task-isolated testing, cross-suite transfer, and search-budget ablations further demonstrate the method's generalization and performance across search budgets. These results establish diffusion-prior texture search as an effective approach that combines closed-loop task disruption with improved scene compatibility in simulated VLA manipulation.

\clearpage
\section*{LLM Usage Statement}
We used \texttt{deepseek-v41-flash} to polish the main-text prose, correct grammatical errors, and provide auxiliary support for experimental work and the organization of experimental content. The authors reviewed and verified the AI-assisted text and experimental content and take full responsibility for the methods, results, and conclusions presented in this paper.

{\small
\bibliographystyle{ieeenat_fullname}
\bibliography{main}}

@article{cohenor2006color,
  author    = {Daniel Cohen-Or and Olga Sorkine and Ran Gal and Tommer Leyvand and Ying-Qing Xu},
  title     = {Color Harmonization},
  journal   = {ACM Transactions on Graphics},
  volume    = {25},
  number    = {3},
  pages     = {624--630},
  year      = {2006},
  doi       = {10.1145/1141911.1141933}
}

@inproceedings{brohan2023rt2visionlanguageactionmodelstransfer,
  author    = {Brianna Zitkovich and Tianhe Yu and Sichun Xu and Peng Xu and Ted Xiao and Fei Xia and Jialin Wu and Paul Wohlhart and Stefan Welker and Ayzaan Wahid and Quan Vuong and Vincent Vanhoucke and Huong Tran and Radu Soricut and Anikait Singh and Jaspiar Singh and Pierre Sermanet and Pannag R. Sanketi and Grecia Salazar and Michael S. Ryoo and Krista Reymann and Kanishka Rao and Karl Pertsch and Igor Mordatch and Henryk Michalewski and Yao Lu and Sergey Levine and Lisa Lee and Tsang-Wei Edward Lee and Isabel Leal and Yuheng Kuang and Dmitry Kalashnikov and Ryan Julian and Nikhil J. Joshi and Alex Irpan and Brian Ichter and Jasmine Hsu and Alexander Herzog and Karol Hausman and Keerthana Gopalakrishnan and Chuyuan Fu and Pete Florence and Chelsea Finn and Kumar Avinava Dubey and Danny Driess and Tianli Ding and Krzysztof Marcin Choromanski and Xi Chen and Yevgen Chebotar and Justice Carbajal and Noah Brown and Anthony Brohan and Montserrat Gonzalez Arenas and Kehang Han},
  title     = {{RT-2}: Vision-Language-Action Models Transfer Web Knowledge to Robotic Control},
  booktitle = {Proceedings of The 7th Conference on Robot Learning},
  series    = {Proceedings of Machine Learning Research},
  volume    = {229},
  pages     = {2165--2183},
  year      = {2023},
  publisher = {PMLR},
  url       = {https://proceedings.mlr.press/v229/zitkovich23a.html}
}

@misc{kim2025finetuningvisionlanguageactionmodels,
  author        = {Moo Jin Kim and Chelsea Finn and Percy Liang},
  title         = {Fine-Tuning Vision-Language-Action Models: Optimizing Speed and Success},
  year          = {2025},
  eprint        = {2502.19645},
  archivePrefix = {arXiv},
  primaryClass  = {cs.RO},
  url           = {https://arxiv.org/abs/2502.19645}
}

@misc{intelligence2025pi05visionlanguageactionmodelopenworld,
  title         = {$\pi_{0.5}$: a Vision-Language-Action Model with Open-World Generalization},
  author        = {{Physical Intelligence} and Kevin Black and Noah Brown and James Darpinian and Karan Dhabalia and Danny Driess and Adnan Esmail and Michael Equi and Chelsea Finn and Niccolo Fusai and Manuel Y. Galliker and Dibya Ghosh and Lachy Groom and Karol Hausman and Brian Ichter and Szymon Jakubczak and Tim Jones and Liyiming Ke and Devin LeBlanc and Sergey Levine and Adrian Li-Bell and Mohith Mothukuri and Suraj Nair and Karl Pertsch and Allen Z. Ren and Lucy Xiaoyang Shi and Laura Smith and Jost Tobias Springenberg and Kyle Stachowicz and James Tanner and Quan Vuong and Homer Walke and Anna Walling and Haohuan Wang and Lili Yu and Ury Zhilinsky},
  year          = {2025},
  eprint        = {2504.16054},
  archivePrefix = {arXiv},
  primaryClass  = {cs.LG},
  url           = {https://arxiv.org/abs/2504.16054}
}

@inproceedings{liu2023liberobenchmarkingknowledgetransfer,
  author    = {Bo Liu and Yifeng Zhu and Chongkai Gao and Yihao Feng and Qiang Liu and Yuke Zhu and Peter Stone},
  title     = {{LIBERO}: Benchmarking Knowledge Transfer for Lifelong Robot Learning},
  booktitle = {Advances in Neural Information Processing Systems},
  volume    = {36},
  pages     = {44776--44791},
  year      = {2023},
  publisher = {Curran Associates, Inc.},
  doi       = {10.52202/075280-1939},
  url       = {https://proceedings.neurips.cc/paper_files/paper/2023/hash/8c3c666820ea055a77726d66fc7d447f-Abstract-Datasets_and_Benchmarks.html}
}

@misc{brown2018adversarialpatch,
  title         = {Adversarial Patch},
  author        = {Tom B. Brown and Dandelion Man{\'e} and Aurko Roy and Mart{\'i}n Abadi and Justin Gilmer},
  year          = {2017},
  eprint        = {1712.09665},
  archivePrefix = {arXiv},
  primaryClass  = {cs.CV},
  url           = {https://arxiv.org/abs/1712.09665},
  note          = {Presented at the NeurIPS 2018 Workshop on Machine Deception}
}

@article{zhang2022adversarialpatchmultiscale,
  author      = {Yichuang Zhang and Yu Zhang and Jiahao Qi and Kangcheng Bin and Hao Wen and Xunqian Tong and Ping Zhong},
  title       = {Adversarial Patch Attack on Multi-Scale Object Detection for UAV Remote Sensing Images},
  journal     = {Remote Sensing},
  volume      = {14},
  number      = {21},
  pages       = {5298},
  year        = {2022},
  doi         = {10.3390/rs14215298},
  url         = {https://doi.org/10.3390/rs14215298},
  institution = {National University of Defense Technology (NUDT)}
}

@inproceedings{chen2023imperceptible,
  author      = {Zihan Chen and Ziyue Wang and Jun-Jie Huang and Wentao Zhao and Xiao Liu and Dejian Guan},
  title       = {Imperceptible Adversarial Attack via Invertible Neural Networks},
  booktitle   = {Proceedings of the AAAI Conference on Artificial Intelligence},
  volume      = {37},
  pages       = {414--424},
  year        = {2023},
  doi         = {10.1609/aaai.v37i1.25115},
  url         = {https://doi.org/10.1609/aaai.v37i1.25115},
  institution = {National University of Defense Technology (NUDT)}
}

@inproceedings{ho2020denoisingdiffusionprobabilisticmodels,
  author    = {Jonathan Ho and Ajay Jain and Pieter Abbeel},
  title     = {Denoising Diffusion Probabilistic Models},
  booktitle = {Advances in Neural Information Processing Systems},
  volume    = {33},
  pages     = {6840--6851},
  year      = {2020},
  publisher = {Curran Associates, Inc.},
  url       = {https://proceedings.neurips.cc/paper_files/paper/2020/hash/4c5bcfec8584af0d967f1ab10179ca4b-Abstract.html}
}

@inproceedings{rombach2022highresolutionimagesynthesislatent,
  author    = {Robin Rombach and Andreas Blattmann and Dominik Lorenz and Patrick Esser and Bj{\"o}rn Ommer},
  title     = {High-Resolution Image Synthesis with Latent Diffusion Models},
  booktitle = {Proceedings of the IEEE/CVF Conference on Computer Vision and Pattern Recognition},
  pages     = {10684--10695},
  year      = {2022},
  doi       = {10.1109/CVPR52688.2022.01042},
  url       = {https://openaccess.thecvf.com/content/CVPR2022/html/Rombach_High-Resolution_Image_Synthesis_With_Latent_Diffusion_Models_CVPR_2022_paper.html}
}

@misc{cheng2025manipulationfacingthreatsevaluating,
  title         = {Manipulation Facing Threats: Evaluating Physical Vulnerabilities in End-to-End Vision Language Action Models},
  author        = {Hao Cheng and Erjia Xiao and Yichi Wang and Chengyuan Yu and Mengshu Sun and Qiang Zhang and Jiahang Cao and Yijie Guo and Ning Liu and Kaidi Xu and Jize Zhang and Chao Shen and Philip Torr and Jindong Gu and Renjing Xu},
  year          = {2024},
  eprint        = {2409.13174},
  archivePrefix = {arXiv},
  primaryClass  = {cs.CV},
  url           = {https://arxiv.org/abs/2409.13174}
}

@inproceedings{wang2025exploringadversarialvulnerabilitiesvisionlanguageaction,
  author    = {Taowen Wang and Cheng Han and James Chenhao Liang and Wenhao Yang and Dongfang Liu and Luna Xinyu Zhang and Qifan Wang and Jiebo Luo and Ruixiang Tang},
  title     = {Exploring the Adversarial Vulnerabilities of Vision-Language-Action Models in Robotics},
  booktitle = {Proceedings of the IEEE/CVF International Conference on Computer Vision},
  pages     = {6948--6958},
  year      = {2025},
  doi       = {10.1109/ICCV51701.2025.00653},
  url       = {https://openaccess.thecvf.com/content/ICCV2025/html/Wang_Exploring_the_Adversarial_Vulnerabilities_of_Vision-Language-Action_Models_in_Robotics_ICCV_2025_paper.html}
}

@misc{xu2025modelagnosticadversarialattackdefense,
  title         = {Model-agnostic Adversarial Attack and Defense for Vision-Language-Action Models},
  author        = {Haochuan Xu and Yun Sing Koh and Shuhuai Huang and Zirun Zhou and Di Wang and Jun Sakuma and Jingfeng Zhang},
  year          = {2025},
  eprint        = {2510.13237},
  archivePrefix = {arXiv},
  primaryClass  = {cs.CV},
  url           = {https://arxiv.org/abs/2510.13237}
}

@misc{lu2026robotsobeypatchuniversal,
  title         = {When Robots Obey the Patch: Universal Transferable Patch Attacks on Vision-Language-Action Models},
  author        = {Hui Lu and Yi Yu and Yiming Yang and Chenyu Yi and Qixin Zhang and Bingquan Shen and Alex C. Kot and Xudong Jiang},
  year          = {2025},
  eprint        = {2511.21192},
  archivePrefix = {arXiv},
  primaryClass  = {cs.CV},
  url           = {https://arxiv.org/abs/2511.21192},
  note          = {Accepted by CVPR 2026}
}

@misc{huang2026traphijackingvlacotreasoning,
  title         = {TRAP: Hijacking VLA CoT-Reasoning via Adversarial Patches},
  author        = {Zhengxian Huang and Wenjun Zhu and Haoxuan Qiu and Xiaoyu Ji and Wenyuan Xu},
  year          = {2026},
  eprint        = {2603.23117},
  archivePrefix = {arXiv},
  primaryClass  = {cs.CR},
  url           = {https://arxiv.org/abs/2603.23117},
  note          = {Accepted by ICML 2026}
}

@misc{fu2026vlahijacktransferablepatchattack,
  title         = {VLA-Hijack: A Transferable Patch Attack against Vision-Language-Action Models via Visual Proprioception Hijacking},
  author        = {Jiyuan Fu and Kaixun Jiang and Jingkai Jia and Zhaoyu Chen and Xueyao Chen and Lingyi Hong and Shuyong Gao and Chenzhi Tan and Dingkang Yang and Wenqiang Zhang},
  year          = {2026},
  eprint        = {2605.28083},
  archivePrefix = {arXiv},
  primaryClass  = {cs.CV},
  url           = {https://arxiv.org/abs/2605.28083}
}

@article{wang2026partiallyobservableadversarialpatch,
  author  = {Xiaofei Wang and Mingliang Han and Tianyu Hao and Yi Yang and Yun-Bo Zhao and Keke Tang},
  title   = {Partially Observable Adversarial Patch Attacks on Vision-Language-Action Models in Robotics},
  journal = {IEEE Robotics and Automation Letters},
  volume  = {11},
  number  = {8},
  pages   = {9351--9358},
  year    = {2026},
  doi     = {10.1109/LRA.2026.3703262},
  url     = {https://doi.org/10.1109/LRA.2026.3703262}
}

@inproceedings{vaswani2017attention,
  author    = {Ashish Vaswani and Noam Shazeer and Niki Parmar and Jakob Uszkoreit and Llion Jones and Aidan N. Gomez and Lukasz Kaiser and Illia Polosukhin},
  title     = {Attention Is All You Need},
  booktitle = {Advances in Neural Information Processing Systems},
  volume    = {30},
  pages     = {5998--6008},
  year      = {2017},
  publisher = {Curran Associates, Inc.},
  url       = {https://proceedings.neurips.cc/paper_files/paper/2017/hash/3f5ee243547dee91fbd053c1c4a845aa-Abstract.html}
}

@inproceedings{dosovitskiy2021image,
  author    = {Alexey Dosovitskiy and Lucas Beyer and Alexander Kolesnikov and Dirk Weissenborn and Xiaohua Zhai and Thomas Unterthiner and Mostafa Dehghani and Matthias Minderer and Georg Heigold and Sylvain Gelly and Jakob Uszkoreit and Neil Houlsby},
  title     = {An Image Is Worth 16x16 Words: Transformers for Image Recognition at Scale},
  booktitle = {International Conference on Learning Representations},
  year      = {2021},
  url       = {https://openreview.net/forum?id=YicbFdNTTy}
}

@inproceedings{radford2021learning,
  author    = {Alec Radford and Jong Wook Kim and Chris Hallacy and Aditya Ramesh and Gabriel Goh and Sandhini Agarwal and Girish Sastry and Amanda Askell and Pamela Mishkin and Jack Clark and Gretchen Krueger and Ilya Sutskever},
  title     = {Learning Transferable Visual Models From Natural Language Supervision},
  booktitle = {Proceedings of the 38th International Conference on Machine Learning},
  series    = {Proceedings of Machine Learning Research},
  volume    = {139},
  pages     = {8748--8763},
  year      = {2021},
  publisher = {PMLR},
  url       = {https://proceedings.mlr.press/v139/radford21a.html}
}

@inproceedings{caron2021emerging,
  author    = {Mathilde Caron and Hugo Touvron and Ishan Misra and Herv{\'e} J{\'e}gou and Julien Mairal and Piotr Bojanowski and Armand Joulin},
  title     = {Emerging Properties in Self-Supervised Vision Transformers},
  booktitle = {Proceedings of the IEEE/CVF International Conference on Computer Vision},
  pages     = {9650--9660},
  year      = {2021},
  doi       = {10.1109/ICCV48922.2021.00951},
  url       = {https://openaccess.thecvf.com/content/ICCV2021/html/Caron_Emerging_Properties_in_Self-Supervised_Vision_Transformers_ICCV_2021_paper.html}
}

@inproceedings{ichter2023do,
  author    = {Brian Ichter and Anthony Brohan and Yevgen Chebotar and Chelsea Finn and Karol Hausman and Alexander Herzog and Daniel Ho and Julian Ibarz and Alex Irpan and Eric Jang and Ryan Julian and Dmitry Kalashnikov and Sergey Levine and Yao Lu and Carolina Parada and Kanishka Rao and Pierre Sermanet and Alexander T. Toshev and Vincent Vanhoucke and Fei Xia and Ted Xiao and Peng Xu and Mengyuan Yan and Noah Brown and Michael Ahn and Omar Cortes and Nicolas Sievers and Clayton Tan and Sichun Xu and Diego Reyes and Jarek Rettinghouse and Jornell Quiambao and Peter Pastor and Linda Luu and Kuang-Huei Lee and Yuheng Kuang and Sally Jesmonth and Nikhil J. Joshi and Kyle Jeffrey and Rosario Jauregui Ruano and Jasmine Hsu and Keerthana Gopalakrishnan and Byron David and Andy Zeng and Chuyuan Kelly Fu},
  title     = {Do As I Can, Not As I Say: Grounding Language in Robotic Affordances},
  booktitle = {Proceedings of The 6th Conference on Robot Learning},
  series    = {Proceedings of Machine Learning Research},
  volume    = {205},
  pages     = {287--318},
  year      = {2023},
  publisher = {PMLR},
  url       = {https://proceedings.mlr.press/v205/ichter23a.html}
}

@inproceedings{brohan2023rt1,
  author    = {Anthony Brohan and Noah Brown and Justice Carbajal and Yevgen Chebotar and Joseph Dabis and Chelsea Finn and Keerthana Gopalakrishnan and Karol Hausman and Alex Herzog and Jasmine Hsu and Julian Ibarz and Brian Ichter and Alex Irpan and Tomas Jackson and Sally Jesmonth and Nikhil J. Joshi and Ryan Julian and Dmitry Kalashnikov and Yuheng Kuang and Isabel Leal and Kuang-Huei Lee and Sergey Levine and Yao Lu and Utsav Malla and Deeksha Manjunath and Igor Mordatch and Ofir Nachum and Carolina Parada and Jodilyn Peralta and Emily Perez and Karl Pertsch and Jornell Quiambao and Kanishka Rao and Michael Ryoo and Grecia Salazar and Pannag Sanketi and Kevin Sayed and Jaspiar Singh and Sumedh Sontakke and Austin Stone and Clayton Tan and Huong Tran and Vincent Vanhoucke and Steve Vega and Quan Vuong and Fei Xia and Ted Xiao and Peng Xu and Sichun Xu and Tianhe Yu and Brianna Zitkovich},
  title     = {{RT-1}: Robotics Transformer for Real-World Control at Scale},
  booktitle = {Robotics: Science and Systems XIX},
  year      = {2023},
  doi       = {10.15607/RSS.2023.XIX.025},
  url       = {https://www.roboticsproceedings.org/rss19/p025.html}
}

@inproceedings{driess2023palme,
  author    = {Danny Driess and Fei Xia and Mehdi S. M. Sajjadi and Corey Lynch and Aakanksha Chowdhery and Brian Ichter and Ayzaan Wahid and Jonathan Tompson and Quan Vuong and Tianhe Yu and Wenlong Huang and Yevgen Chebotar and Pierre Sermanet and Daniel Duckworth and Sergey Levine and Vincent Vanhoucke and Karol Hausman and Marc Toussaint and Klaus Greff and Andy Zeng and Igor Mordatch and Pete Florence},
  title     = {{PaLM-E}: An Embodied Multimodal Language Model},
  booktitle = {Proceedings of the 40th International Conference on Machine Learning},
  series    = {Proceedings of Machine Learning Research},
  volume    = {202},
  pages     = {8469--8488},
  year      = {2023},
  publisher = {PMLR},
  url       = {https://proceedings.mlr.press/v202/driess23a.html}
}

@inproceedings{jiang2023vima,
  author    = {Yunfan Jiang and Agrim Gupta and Zichen Zhang and Guanzhi Wang and Yongqiang Dou and Yanjun Chen and Li Fei-Fei and Anima Anandkumar and Yuke Zhu and Linxi Fan},
  title     = {{VIMA}: Robot Manipulation with Multimodal Prompts},
  booktitle = {Proceedings of the 40th International Conference on Machine Learning},
  series    = {Proceedings of Machine Learning Research},
  volume    = {202},
  pages     = {14975--15022},
  year      = {2023},
  publisher = {PMLR},
  url       = {https://proceedings.mlr.press/v202/jiang23b.html}
}

@inproceedings{shridhar2023perceiver,
  author    = {Mohit Shridhar and Lucas Manuelli and Dieter Fox},
  title     = {Perceiver-Actor: A Multi-Task Transformer for Robotic Manipulation},
  booktitle = {Proceedings of The 6th Conference on Robot Learning},
  series    = {Proceedings of Machine Learning Research},
  volume    = {205},
  pages     = {785--799},
  year      = {2023},
  publisher = {PMLR},
  url       = {https://proceedings.mlr.press/v205/shridhar23a.html}
}

@inproceedings{chi2023diffusion,
  author    = {Cheng Chi and Siyuan Feng and Yilun Du and Zhenjia Xu and Eric Cousineau and Benjamin Burchfiel and Shuran Song},
  title     = {Diffusion Policy: Visuomotor Policy Learning via Action Diffusion},
  booktitle = {Robotics: Science and Systems XIX},
  year      = {2023},
  doi       = {10.15607/RSS.2023.XIX.026},
  url       = {https://www.roboticsproceedings.org/rss19/p026.html}
}

@inproceedings{goodfellow2015explaining,
  author    = {Ian J. Goodfellow and Jonathon Shlens and Christian Szegedy},
  title     = {Explaining and Harnessing Adversarial Examples},
  booktitle = {International Conference on Learning Representations},
  year      = {2015},
  url       = {https://arxiv.org/abs/1412.6572}
}

@inproceedings{carlini2017towards,
  author    = {Nicholas Carlini and David Wagner},
  title     = {Towards Evaluating the Robustness of Neural Networks},
  booktitle = {Proceedings of the IEEE Symposium on Security and Privacy},
  pages     = {39--57},
  year      = {2017},
  doi       = {10.1109/SP.2017.49},
  url       = {https://doi.org/10.1109/SP.2017.49}
}

@inproceedings{eykholt2018robust,
  author    = {Kevin Eykholt and Ivan Evtimov and Earlence Fernandes and Bo Li and Amir Rahmati and Chaowei Xiao and Atul Prakash and Tadayoshi Kohno and Dawn Song},
  title     = {Robust Physical-World Attacks on Deep Learning Visual Classification},
  booktitle = {Proceedings of the IEEE/CVF Conference on Computer Vision and Pattern Recognition},
  pages     = {1625--1634},
  year      = {2018},
  doi       = {10.1109/CVPR.2018.00175},
  url       = {https://openaccess.thecvf.com/content_cvpr_2018/html/Eykholt_Robust_Physical-World_Attacks_CVPR_2018_paper.html}
}

@inproceedings{athalye2018synthesizing,
  author    = {Anish Athalye and Logan Engstrom and Andrew Ilyas and Kevin Kwok},
  title     = {Synthesizing Robust Adversarial Examples},
  booktitle = {Proceedings of the 35th International Conference on Machine Learning},
  series    = {Proceedings of Machine Learning Research},
  volume    = {80},
  pages     = {284--293},
  year      = {2018},
  publisher = {PMLR},
  url       = {https://proceedings.mlr.press/v80/athalye18b.html}
}

@inproceedings{madry2018towards,
  author    = {Aleksander Madry and Aleksandar Makelov and Ludwig Schmidt and Dimitris Tsipras and Adrian Vladu},
  title     = {Towards Deep Learning Models Resistant to Adversarial Attacks},
  booktitle = {International Conference on Learning Representations},
  year      = {2018},
  url       = {https://openreview.net/forum?id=rJzIBfZAb}
}

@inproceedings{hendrycks2019benchmarking,
  author    = {Dan Hendrycks and Thomas G. Dietterich},
  title     = {Benchmarking Neural Network Robustness to Common Corruptions and Perturbations},
  booktitle = {International Conference on Learning Representations},
  year      = {2019},
  url       = {https://openreview.net/forum?id=HJz6tiCqYm}
}

@article{gretton2012kernel,
  author  = {Arthur Gretton and Karsten M. Borgwardt and Malte J. Rasch and Bernhard Sch{\"o}lkopf and Alexander Smola},
  title   = {A Kernel Two-Sample Test},
  journal = {Journal of Machine Learning Research},
  volume  = {13},
  number  = {25},
  pages   = {723--773},
  year    = {2012},
  url     = {https://jmlr.org/papers/v13/gretton12a.html}
}

@inproceedings{sohldickstein2015deep,
  author    = {Jascha Sohl-Dickstein and Eric A. Weiss and Niru Maheswaranathan and Surya Ganguli},
  title     = {Deep Unsupervised Learning Using Nonequilibrium Thermodynamics},
  booktitle = {Proceedings of the 32nd International Conference on Machine Learning},
  series    = {Proceedings of Machine Learning Research},
  volume    = {37},
  pages     = {2256--2265},
  year      = {2015},
  publisher = {PMLR},
  url       = {https://proceedings.mlr.press/v37/sohl-dickstein15.html}
}

@inproceedings{song2021score,
  author    = {Yang Song and Jascha Sohl-Dickstein and Diederik P. Kingma and Abhishek Kumar and Stefano Ermon and Ben Poole},
  title     = {Score-Based Generative Modeling through Stochastic Differential Equations},
  booktitle = {International Conference on Learning Representations},
  year      = {2021},
  url       = {https://openreview.net/forum?id=PxTIG12RRHS}
}

@inproceedings{nichol2021improved,
  author    = {Alexander Quinn Nichol and Prafulla Dhariwal},
  title     = {Improved Denoising Diffusion Probabilistic Models},
  booktitle = {Proceedings of the 38th International Conference on Machine Learning},
  series    = {Proceedings of Machine Learning Research},
  volume    = {139},
  pages     = {8162--8171},
  year      = {2021},
  publisher = {PMLR},
  url       = {https://proceedings.mlr.press/v139/nichol21a.html}
}

@inproceedings{meng2022sdedit,
  author    = {Chenlin Meng and Yutong He and Yang Song and Jiaming Song and Jiajun Wu and Jun-Yan Zhu and Stefano Ermon},
  title     = {{SDEdit}: Guided Image Synthesis and Editing with Stochastic Differential Equations},
  booktitle = {International Conference on Learning Representations},
  year      = {2022},
  url       = {https://openreview.net/forum?id=aBsCjcPu_tE}
}

@inproceedings{karras2022elucidating,
  author    = {Tero Karras and Miika Aittala and Timo Aila and Samuli Laine},
  title     = {Elucidating the Design Space of Diffusion-Based Generative Models},
  booktitle = {Advances in Neural Information Processing Systems},
  volume    = {35},
  pages     = {26565--26577},
  year      = {2022},
  publisher = {Curran Associates, Inc.},
  url       = {https://proceedings.neurips.cc/paper_files/paper/2022/hash/a98846e9d9cc01cfb87eb694d946ce6b-Abstract-Conference.html}
}

@inproceedings{croce2020reliable,
  author    = {Francesco Croce and Matthias Hein},
  title     = {Reliable Evaluation of Adversarial Robustness with an Ensemble of Diverse Parameter-Free Attacks},
  booktitle = {Proceedings of the 37th International Conference on Machine Learning},
  series    = {Proceedings of Machine Learning Research},
  volume    = {119},
  pages     = {2206--2216},
  year      = {2020},
  publisher = {PMLR},
  url       = {https://proceedings.mlr.press/v119/croce20b.html}
}

@inproceedings{karmon2018lavan,
  author    = {Danny Karmon and Daniel Zoran and Yoav Goldberg},
  title     = {{LaVAN}: Localized and Visible Adversarial Noise},
  booktitle = {Proceedings of the 35th International Conference on Machine Learning},
  series    = {Proceedings of Machine Learning Research},
  volume    = {80},
  pages     = {2507--2515},
  year      = {2018},
  publisher = {PMLR},
  url       = {https://proceedings.mlr.press/v80/karmon18a.html}
}

@inproceedings{thys2019fooling,
  author    = {Simen Thys and Wiebe Van Ranst and Toon Goedem{\'e}},
  title     = {Fooling Automated Surveillance Cameras: Adversarial Patches to Attack Person Detection},
  booktitle = {Proceedings of the IEEE/CVF Conference on Computer Vision and Pattern Recognition Workshops},
  pages     = {49--55},
  year      = {2019},
  doi       = {10.1109/CVPRW.2019.00012},
  url       = {https://openaccess.thecvf.com/content_CVPRW_2019/html/CV-COPS/Thys_Fooling_Automated_Surveillance_Cameras_Adversarial_Patches_to_Attack_Person_Detection_CVPRW_2019_paper.html}
}

@inproceedings{hu2021naturalistic,
  author    = {Yu-Chih-Tuan Hu and Bo-Han Kung and Daniel Stanley Tan and Jun-Cheng Chen and Kai-Lung Hua and Wen-Huang Cheng},
  title     = {Naturalistic Physical Adversarial Patch for Object Detectors},
  booktitle = {Proceedings of the IEEE/CVF International Conference on Computer Vision},
  pages     = {7828--7837},
  year      = {2021},
  doi       = {10.1109/ICCV48922.2021.00775}
}

@inproceedings{liang2023advdm,
  author    = {Liang, Chumeng and Wu, Xiaoyu and Hua, Yang and Zhang, Jiaru and Xue, Yiming and Song, Tao and Xue, Zhengui and Ma, Ruhui and Guan, Haibing},
  title     = {Adversarial Example Does Good: Preventing Painting Imitation from Diffusion Models via Adversarial Examples},
  booktitle = {Proceedings of the 40th International Conference on Machine Learning},
  series    = {Proceedings of Machine Learning Research},
  volume    = {202},
  pages     = {20763--20786},
  publisher = {PMLR},
  year      = {2023},
  url       = {https://proceedings.mlr.press/v202/liang23g.html}
}

@inproceedings{chen2023diffattack,
  author    = {Kang, Mintong and Song, Dawn and Li, Bo},
  title     = {DiffAttack: Evasion Attacks Against Diffusion-Based Adversarial Purification},
  booktitle = {Advances in Neural Information Processing Systems 36},
  year      = {2023},
  eprint    = {2311.16124},
  archivePrefix = {arXiv},
  url       = {https://papers.nips.cc/paper_files/paper/2023/hash/ea0b28cbbd0cbc45ec4ac38e92da9cb2-Abstract-Conference.html}
}

@inproceedings{hendrycks2021natural,
  author    = {Hendrycks, Dan and Zhao, Kevin and Basart, Steven and Steinhardt, Jacob and Song, Dawn},
  title     = {Natural Adversarial Examples},
  booktitle = {Proceedings of the IEEE/CVF Conference on Computer Vision and Pattern Recognition},
  pages     = {15262--15271},
  year      = {2021},
  doi       = {10.1109/CVPR46437.2021.01501}
}

@inproceedings{luo2024visual,
  author    = {Qi, Xiangyu and Huang, Kaixuan and Panda, Ashwinee and Henderson, Peter and Wang, Mengdi and Mittal, Prateek},
  title     = {Visual Adversarial Examples Jailbreak Aligned Large Language Models},
  booktitle = {Proceedings of the 38th AAAI Conference on Artificial Intelligence},
  volume    = {38},
  number    = {19},
  pages     = {21527--21536},
  year      = {2024},
  doi       = {10.1609/aaai.v38i19.30150}
}

@inproceedings{gong2023figstep,
  author        = {Gong, Yichen and Ran, Delong and Liu, Jinyuan and Wang, Conglei and Cong, Tianshuo and Wang, Anyu and Duan, Sisi and Wang, Xiaoyun},
  title         = {FigStep: Jailbreaking Large Vision-Language Models via Typographic Visual Prompts},
  booktitle     = {Proceedings of the 39th AAAI Conference on Artificial Intelligence},
  volume        = {39},
  number        = {22},
  pages         = {23951--23959},
  year          = {2025},
  doi           = {10.1609/aaai.v39i22.34568},
  eprint        = {2311.05608},
  archivePrefix = {arXiv},
  primaryClass  = {cs.CR}
}

@inproceedings{bailey2024imagehijacks,
  author    = {Bailey, Luke and Ong, Euan and Russell, Stuart and Emmons, Scott},
  title     = {Image Hijacks: Adversarial Images can Control Generative Models at Runtime},
  booktitle = {Proceedings of the 41st International Conference on Machine Learning},
  series    = {Proceedings of Machine Learning Research},
  volume    = {235},
  pages     = {2443--2455},
  publisher = {PMLR},
  year      = {2024}
}
\end{document}